%% file: acl_latex.tex
\documentclass[11pt]{article}

\usepackage[final]{acl}

\usepackage{times}
\usepackage{latexsym}
\usepackage[T1]{fontenc}
\usepackage[utf8]{inputenc}
\usepackage{microtype}
\usepackage{inconsolata}
\usepackage{graphicx}
\usepackage{booktabs}
\usepackage{multirow}
\usepackage{amsmath}
\usepackage{enumitem}

\graphicspath{{figures/}}

\newif\iflcfm \lcfmfalse

\title{When Dynamic Scaling Fails: A Scale--State Feedback Governs\\
Recurrent-State Quantization in Linear-Attention and SSM LLMs}

\author{Yuhang Jiang \\
  Independent Researcher \\
  \texttt{jyhtjtj@gmail.com}}

\begin{document}
\maketitle

\begin{abstract}
\input{sec_abstract}
\end{abstract}

\input{sec_intro}
\input{sec_related}
\input{sec_setup}
\input{sec_frontier}
\input{sec_mechanism}
\input{sec_policy}
\input{sec_whatworks}
\input{sec_survival}
\input{sec_discussion}
\input{sec_backmatter}

\bibliography{custom}

\appendix
\input{sec_appendix}

\end{document}

%% file: sec_abstract.tex
Linear-attention and state-space LLMs carry long context in a fixed-size recurrent state, and quantizing that
state is the natural way to cut its cost during long decode. Unlike a weight tensor, the state is rewritten at
every step, so a quantizer applied at each step sits inside the recurrence and its errors compound. INT4 that
leaves synthetic recall perfectly intact at the start destroys it a thousand steps later. The mechanism is a feedback between the quantizer scale and the state
it is derived from. Yoking that scale to a full-precision reference removes the resulting norm explosion
entirely, yet recovers only 1.8 of the 5.4 nats lost, so numerical instability and functional degradation are
separable failures. The remainder is resolution-dependent, and in our per-step setting every
outlier-aware scheme we test loses to plain persistent scaling. Allocating bits across layers does help, but under
article-disjoint splits a learned allocation collapses onto the trivial rule of protecting the first few
layers. Run out to 32K, group-wise INT4 is worth no more
than discarding the state. Contain the scale rather than locate outliers, and report the
decode horizon at which a bit-width was certified.

%% file: sec_intro.tex
\section{Introduction}
\label{sec:intro}

Suppose you quantize the recurrent state of a linear-attention model to INT4 and check it the usual way, with
needle-in-a-haystack retrieval. Recall comes back perfect at every length you measure, so you ship it. Past the
longest length you tested, the best of the four-bit policies still recovers the planted needle on demand, and its
perplexity on ordinary text has risen eightfold. It remembers the fact and has stopped being a language model.

\begin{figure*}[t]
\centering
\includegraphics[width=\textwidth]{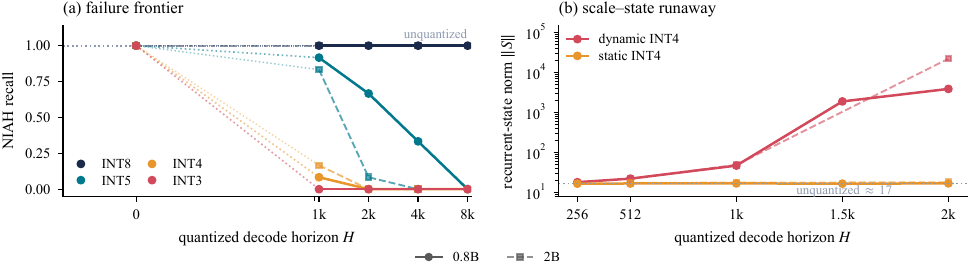}
\caption{\textbf{The phenomenon.} (a) NIAH recall over 12 seeds against quantized decode horizon $H$. Every
bit-width is intact at $H=0$ and decays as the state is overwritten; INT5 traces the frontier. (b) Mean
recurrent-state norm on independent articles, log scale: per-token dynamic INT4 inflates it by two to three
orders of magnitude, while the static curves and the unquantized reference coincide along the bottom. Nothing is
measured between $H=0$ and $H=1024$, so that span is dotted; 2B was not run at $H=8192$.}
\label{fig:frontier}
\end{figure*}

Both halves of that story are measured below, and both are invisible to the way recurrent-state quantization is
currently reported. Linear-attention and state-space models replace the growing KV cache of softmax attention
with a \emph{fixed-size recurrent state}: Gated DeltaNet (GDN) powers the linear layers of Qwen3-Next
\citep{qwen3next}, Kimi Linear \citep{kimilinear} builds on it, and Mamba-2 is a widely
used SSM. That state is a real deployment cost during long decode, so quantizing it is
the natural remedy. But a recurrent state is not a weight tensor. It is \emph{rewritten at every decode step}, so
a quantizer applied at each step sits inside the recurrence and its errors compound, and \textbf{a bit-width
certified as safe at short exposure need not stay safe as decode proceeds}. Per-token dynamic INT4 leaves
synthetic recall perfectly intact at $H=0$ and destroys it by $H=1024$ (Figure~\ref{fig:frontier}a).

Three things have kept that axis out of view. The closest developed literature is low-bit KV-cache quantization
\citep{hooper2024kvquant,liu2024kivi}, where the cache is stored rather than recurrent, so error never re-enters
the state and exposure is not reported as a variable. Quantization research more broadly grew up on weights and
activations \citep{dettmers2022llmint8,xiao2023smoothquant,frantar2023gptq}, neither of which carries its own
quantization error into the next step. And the probe everyone reaches for is retrieval accuracy, which turns out to be the least sensitive
instrument available for this failure. The eightfold rise in the opening paragraph is what that insensitivity
costs, and a warning in its own right.

Prior work quantizes these caches with architecture-specific machinery: Q-Mamba \citep{qmamba} reaches an INT4
Mamba-2 state cache via trained DSQ/ESR alongside weights and activations, and the Nemotron~3 line ships FP16
with stochastic rounding for the Mamba SSM cache \citep{nemotron3ultra}. We instead study \emph{state-only,
training-free fake-quantization}, to isolate what the recurrence does to quantization error independently of any
trained corrector.

We contribute:
\begin{itemize}[leftmargin=*,itemsep=1pt,topsep=1pt]
\item A \textbf{failure frontier} in (bit-width $\times$ decode horizon), making decode horizon an explicit
stability axis for recurrent-cache quantization (\S\ref{sec:frontier}).
\item A \textbf{scale--state feedback mechanism} for per-token dynamic scaling, causally identified by a
scale-yoking intervention that removes the numerical runaway but only part of the functional loss, and a
unifying account of \emph{containment} that explains why a calibrated ceiling, periodic reset and persistent
scaling all work, and why which of them works best depends on the architecture (\S\ref{sec:mech},
\S\ref{sec:contain}).
\item A \textbf{universal policy$\times$exposure interaction} with a \textbf{model-dependent} short-horizon
reversal, replicated on independent articles across architectures and scales (\S\ref{sec:policy}).
\item A \textbf{negative result}: under per-step quantization, every outlier-locating allocation we test loses
to plain group-wise scaling, while
per-layer allocation helps only to collapse onto a trivial rule (\S\ref{sec:whatworks}), and a \textbf{survival
analysis} showing that short evaluations certify policies as safe long before their failures appear
(\S\ref{sec:survival}).
\end{itemize}

%% file: sec_related.tex
\section{Related work}
\label{sec:related}

\paragraph{Linear-attention and state-space LLMs.} Linear attention replaces softmax attention with a
kernel-feature recurrence and a fixed-size state \citep{katharopoulos2020linear}; the delta rule sharpens that
recurrence \citep{schlag2021deltanet,yang2024deltanet}, and Gated DeltaNet \citep{yang2024gdn} is the linear
component of Qwen3-Next \citep{qwen3next}, while Kimi Linear \citep{kimilinear} extends it with finer-grained
gating. State-space models \citep{gu2023mamba,dao2024mamba2}
carry context in an SSM state. All of them share the property we study: a fixed-size recurrent state, overwritten
each step.

\paragraph{Cache quantization.} For softmax attention the analogous target is the growing KV cache, and low-bit
KV-cache quantization is well developed \citep{hooper2024kvquant,liu2024kivi}. The recurrent case differs: the
cache is constant-size but \emph{recurrent}, so each step's error enters the state that produces the next one.
Beyond Q-Mamba \citep{qmamba}, Quamba2 \citep{quamba2} quantizes Mamba weights and activations (W8A8 through
W4A16) together with the inputs of the linear recurrence, MambaQuant \citep{mambaquant} applies variance-aligning
rotations to weights and activations, and ReplaySSM \citep{replayssm} recomputes state from cached
\emph{inputs} rather than quantizing it at all.

Of these only Q-Mamba reaches the persistent state, and only alongside weights and activations. Concurrent work
DAMP \citep{damp} quantizes the GDN and KDA state directly, choosing high-risk channels offline by
quantization-error energy and decay persistence. It runs at about 9.9 bits per state value on 35B and 48B models,
with deployment kernels, and reports that INT4 and NVFP4 collapse accuracy. We hold everything else in full
precision, work below five bits, and vary how long the state is quantized, an axis DAMP holds fixed. \citet{minima} instead quantize weights and activations of a hybrid 27B model and find injected
noise flat over 32K tokens. There the stored state stays in high precision, so the scale never becomes a function
of the corrupted state.

\iflcfm
\else
Three further concurrent works quantize the recurrent state directly, and each improves the quantizer where we
hold it fixed. \citet{li2026preserving} trace the per-step error to updates smaller than the grid spacing,
which are lost in rounding and lost coherently, and recover them with a low-rank bypass and
direction-preserving rounding. Bits can instead be allocated per channel, weighted by the observability
Gramian and by decay and range \citep{chen2026lowbit}. LeapQuant quantizes once per window, buffers the
intermediate updates in high precision, and keeps outliers as high-precision compensator tokens, with
deployment kernels \citep{pan2026leapquant}. We vary a plain per-step quantizer's scale policy against
quantized exposure, and compare against none of them.
\fi

\paragraph{Outliers and group-wise quantization.} A large body of LLM-quantization work targets activation
outliers---mixed-precision on outlier channels \citep{dettmers2022llmint8}, migrating outlier scale into weights
\citep{xiao2023smoothquant}, second-order error compensation \citep{frantar2023gptq}, and activation-aware
protection of salient weights \citep{lin2023awq}---and group-wise scaling is a standard remedy.
In our per-step setting, \emph{locating} the outlier channels does not beat
uniform group-wise quantization, which we read as the failure being driven by a scale--state feedback to be
\emph{contained} rather than by a static set of outliers. That is a statement about per-step requantization:
under per-window quantization, keeping outliers in high precision does help \citep{pan2026leapquant}. Sensitivity-aware mixed precision is likewise well established, from Hessian-trace layer allocation
\citep{hawqv2} to layer-wise KV precision search \citep{kvtuner} and per-head rate-distortion allocation
\citep{ratequant}. Our leak-free audit (\S\ref{sec:whatworks}) shows that, at layer granularity, this
machinery adds nothing here---we do not test within-layer, decay-aware allocation
\citep{damp,chen2026lowbit}---since sensitivity is so concentrated in the first layers that a learned allocator, a knapsack
optimizer, and a trivial protect-first-few-layers baseline coincide.

%% file: sec_setup.tex
\section{Setup}
\label{sec:setup}

\paragraph{Models.} Qwen3.5-0.8B and Qwen3.5-2B (official Gated DeltaNet hybrid \citep{yang2024gdn,qwen35},
262K context, BF16 compute with FP32 recurrent state) and Mamba-2-1.3B \citep{dao2024mamba2}
(\texttt{state-spaces}, native loader). The GDN recurrent state has shape
$(1,16,128,128)$; the Mamba SSM state $(1,64,64,128)$.

\paragraph{Quantization.} We apply symmetric per-channel INT-$b$ fake-quantization to the recurrent state at
each decode step: round, clamp, dequantize. Scaling at channel or vector granularity is the standard response to
activation outliers \citep{dettmers2022llmint8,xiao2023smoothquant}, and group-wise scaling the one KV-cache
quantizers use \citep{liu2024kivi,hooper2024kvquant}; we carry both over to the recurrent state unchanged. Of the three scaling \emph{policies} we compare, \textbf{dynamic}
takes the per-step absmax, $s_{t,c}=\max|S_{t,c}|/q_{\max}$. \textbf{Static} (persistent) uses one offline-calibrated
fixed scale, collected over multiple held-out sequences and timepoints and chosen to be constrained-MSE-optimal.
\textbf{Group-wise} is persistent with a separate scale for each group of $G=32$ state channels, at
$5.0$ bits per element against $4.0$ for the per-channel policies.

\paragraph{Tasks and protocol.} (a) \emph{NIAH recall} with fictional needles, in the
needle-in-a-haystack family \citep{hsieh2024ruler} (paired McNemar). (b)
\emph{Natural long-document continuation}: process $H$ real tokens with per-step quantization, then teacher-force
256 natural tokens and measure per-document excess negative log-likelihood (NLL) against the
unquantized model. A \textbf{zero-state control} confirms each task
uses the recurrent state, and calibration and test are \textbf{article-disjoint} (independent
WikiText-103 articles \citep{merity2017wikitext}). For the exposure analysis we fix total context to 2048 and vary only the \emph{quantized
exposure} $H$: the last $H$ steps are quantized, the first $2048-H$ are an unquantized warm-up, so every $H$
predicts the \emph{same} continuation.

\paragraph{Statistics.} All comparisons are paired by document or by seed, and effects are reported with $t$ and
discordant counts. The exposure$\times$policy interaction is a difference-in-differences. Sample sizes differ by
experiment and are given with each result: 12 or 16 seeds for the recall probes, 16 documents for the
continuation and containment experiments, 18 held-out articles for the allocation audit, and 30 for the survival
analysis.

%% file: sec_frontier.tex
\section{The failure frontier}
\label{sec:frontier}

The usual way to report a state quantizer is to name a bit-width. Sweeping bit-width against decode horizon shows
what that leaves out (Figure~\ref{fig:frontier}a; exact recall values in Appendix
Table~\ref{tab:frontier}). On GDN, INT8 is safe at every horizon measured, INT5 degrades gradually, INT4 collapses, and INT3 fails
immediately, and the pattern is orthogonal to model scale, replicating at 0.8B and 2B. Mamba-2-1.3B has the same
frontier at a longer length scale: INT4 recall is intact at $H=1024$, where GDN-0.8B is already at $0.08$, and
falls to $0.33$ by $H=2048$; INT8 holds through $H=2048$ and falls to $0.08$ at $H=4096$, where the unquantized
model still answers $0.75$. At $H=8192$ nothing answers, full precision included, which brackets the SSM
frontier without tracing it. That is safety as the recall probe sees it. On natural continuation the same INT8 state costs $0.26$ nats at
$H=2048$ and $0.62$ at $H=32768$, a growth of $0.35$ paired across contexts ($t=10.9$), and its perplexity
penalty rises from $1.30\times$ the unquantized model to $1.85\times$ (Appendix~\ref{sec:appendix-32k}). The
bit-width that survives a needle test is still losing language-modelling quality as decode proceeds.
Every configuration is horizon-dependent too. Recall at $H=0$ is always intact, because the state has not yet
been overwritten under quantization, and it decays as more decode steps pass through the quantizer. Lower
bit-width is worse, longer horizon is worse, and the two axes are not separable. A bit-width therefore has to be reported together with the horizon at which it was measured, and this much holds
independently of any mechanistic claim.

%% file: sec_mechanism.tex
\section{Mechanism: a scale--state feedback loop}
\label{sec:mech}

The frontier says when INT4 breaks down, not why. The natural suspect is the part of the pipeline that looks most
defensible. Per-token dynamic quantization re-derives the per-channel step size from the current state at every
step, $s_{t,c}=\max|S_{t,c}|/q_{\max}$, which is exactly what one would do to track a moving distribution and
is how KIVI scales the value cache \citep{liu2024kivi}. Inside
a recurrence it also couples the quantizer to its own output, and that coupling turns out to be destabilizing.

\begin{figure*}[t]
\centering
\includegraphics[width=\textwidth]{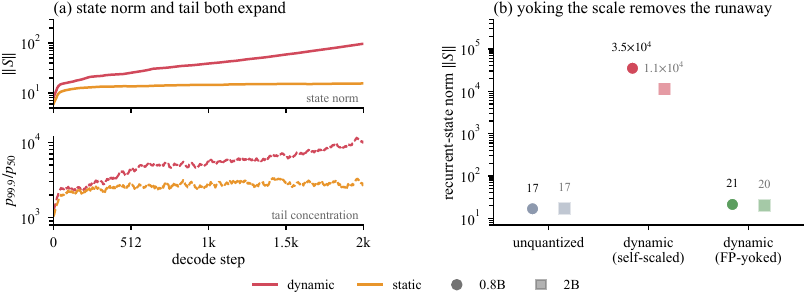}
\caption{\textbf{The feedback, and cutting it.} (a) Single-layer decode trajectory, boxcar smoothed. Norm and
tail concentration get separate axes because they carry different units; both expand under dynamic INT4 and stay
bounded under a persistent scale. (b) Taking the quantizer step size from the FP state instead of the quantized
one removes the explosion at both scales. The three columns are independent arms, not a time course.}
\label{fig:mechanism}
\end{figure*}

\paragraph{Numerical runaway is real and policy-driven.} Under per-token dynamic INT4, the closed-loop
recurrent-state norm grows super-linearly with decode length (Figure~\ref{fig:frontier}b). On independent
articles at $H=2048$, the GDN state norm inflates from its unquantized value of $\sim 17$ to
$3.9\times10^{3}$ at 0.8B and $2.2\times10^{4}$ at 2B. A persistent fixed-range static quantizer holds the same
norm at $\sim 17$--$18$, and a low clip rate is enough to do it: static scaling that clips 0.02\% of elements
already prevents the explosion. Mamba-2-1.3B differs in degree: the same policy grows the SSM-state norm
only from $21.8$ to $39$--$42$ across runs, a factor of $1.8$--$1.9$, yet function still collapses at long
horizon (\S\ref{sec:policy}). The numerical severity is architecture-dependent; the functional failure is not.

\paragraph{What the loop looks like step by step.} Within a dynamic decode, three quantities co-expand: the
per-channel scale, the tail concentration $p_{99.9}/p_{50}$ of $|S|$, and the state norm. In a representative
layer the tail ratio runs from $1.2\times10^{3}$ to $6.1\times10^{3}$ and the norm from $2.8$ to $98$
(Figure~\ref{fig:mechanism}a), whereas static quantization keeps all three bounded. That is what an outlier
enlarging the scale, coarsening the bulk resolution, and admitting further error would look like, but the
trajectory evidence is only correlational. The scale-vs-max relation is partly definitional, a bulk-resolution
proxy did \emph{not} monotonically collapse, and $p_{99.9}/p_{50}$ is unstable on near-sparse states. The loop is
the cause of the runaway, and is consistent with, but not proven to be, the whole path to functional loss.

\paragraph{Causal identification via scale-yoking.} We cut the feedback edge directly. Running the FP and a
per-token-dynamic-INT4 model in lockstep on the same tokens, we \emph{yoke} the quantizer step at each layer to
the FP state, $s_t=\max|S_t^{\mathrm{FP}}|/q_{\max}$. The state is still quantized at every step, but the scale
no longer depends on the runaway quantized state. Yoking \textbf{eliminates the numerical explosion} (Figure~\ref{fig:mechanism}b). At $H=2048$ the mean
GDN-0.8B state norm falls from a self-scaled $3.5\times10^{4}$ to $21$ under yoking, against an unquantized
$17$. That mean is driven by 5 of 16 articles above $10^{3}$; the median tells the same story less
dramatically, $268$ self-scaled against $21$ and $17$. The 2B result has the same shape,
$1.1\times10^{4}$ falling to $20$.
The self-scaled mean is heavy-tailed and spans two orders of magnitude across runs, while excess NLL for the same condition varies by under $2\%$, so only its order of magnitude should be
read as a finding (Appendix~\ref{sec:appendix-proto}).

The functional recovery is far smaller than the numerical one. Yoking returns $1.82$ nats of excess NLL at 0.8B,
from $5.39$ to $3.57$ (paired $t=-13.1$), and $1.35$ nats at 2B ($t=-6.9$). Two thirds of the damage survives an
intervention that removes the runaway completely. The two failures are separable, and whatever accounts for the
residual is not a stability problem (\S\ref{sec:whatworks}).

\section{What contains the loop}
\label{sec:contain}

The numerical half, at least, is containable, and which remedies work says what the loop is made of.

\paragraph{A ceiling on the scale, not a hunt for outliers.} Outliers drive the scale in weight and
activation quantization, and the methods that work there isolate or rescale them
\citep{dettmers2022llmint8,xiao2023smoothquant,lin2023awq}. Two candidate remedies can be separated with
matched-rate controls (Table~\ref{tab:contain}). The first caps the dynamic scale at a ceiling $c$ taken from
held-out calibration, $s=\min(\max|S|/q_{\max},\,c)$. Adding exact preservation of the overflowing elements on
top of that cap changes nothing, moving excess NLL by $0.18$ nats ($t=1.53$): keeping the outliers exactly does
not help, so it is the scale ceiling that matters. The second remedy interrupts the loop in time. Zeroing the recurrent state every $K$ steps recovers about two
nats of the dynamic runaway at $H=2048$ (Table~\ref{tab:contain}), and the benefit is insensitive to
$K\in\{128,256,512\}$.
A persistent calibrated scale goes further and removes the edge: a scale that never reads the state cannot be
driven by it. Capping, periodic reset, and persistent calibration bound, interrupt, and
eliminate the same edge, which is why the methods of \S\ref{sec:whatworks} that instead \emph{locate} outliers
fail.

\paragraph{Numerical containment $\neq$ functional preservation.} Bounding the norm does not by itself preserve
function. Static-INT5 keeps the norm near $17$ across horizons and still incurs excess NLL ranging from $1.75$ to
$5.1$ across conditions. Bounded norm does not imply bounded functional loss, and a static policy is not a
constant NLL bias.

\paragraph{Removing the feedback is necessary; one scale is not always enough.} Among the policies that remove
the edge, how many scales they spend separates the architectures (Table~\ref{tab:cross}). On both GDN models a
single persistent scale is worse than discarding the recurrent state outright, by $1.77$ nats at 0.8B
($t=15.2$, no article better) and indistinguishably at 2B ($t=0.04$), while persistent group-wise INT4, which
spends four scales where static spends one, beats the control by $0.99$ and $1.93$ nats. On Mamba-2 the
ordering reverses: one scale already beats discarding the state ($t=-5.00$, 16 of 16 articles), and the extra
scales cost $0.16$ nats ($t=2.39$) while raising the modeled rate from $4.0$ to $5.0$ bits per element
(\S\ref{sec:setup}). Per-token dynamic INT4 is worse than discarding the state in all three
models. The feedback behaves the same way in both architectures. How many scales it then takes to leave a
state worth keeping does not transfer.

\begin{table}[t]
\centering\small\setlength{\tabcolsep}{3.4pt}
\begin{tabular}{lcc}
\toprule
Intervention ($H=2048$) & excess NLL & paired $t$ \\
\midrule
dynamic INT4              & 5.47 & --- \\
periodic reset ($K=128$) & 3.38 & $t=-6.85$ \\
capped dynamic           & 3.58 & $t=-11.0$ \\
\;\;+ exact escape        & 3.40 & $t=-1.53$ \\
group-wise INT4          & \textbf{2.09} & $t=-13.4$ \\
\bottomrule
\end{tabular}
\caption{Containment beats location. All three containment interventions reduce the dynamic runaway; adding exact
outlier escape on top of a scale cap does not help ($\approx$ capped), so scale decoupling---not outlier
removal---is the operative mechanism. Paired $t$ is against dynamic INT4, except for the indented row,
which is against capped dynamic. GDN-0.8B, natural continuation, $n=16$ documents, single batch.}
\label{tab:contain}
\end{table}

\begin{table*}[t]
\centering\small\setlength{\tabcolsep}{6pt}
\begin{tabular}{lccc}
\toprule
Excess NLL at $H=2048$ & GDN-0.8B & GDN-2B & Mamba-2-1.3B \\
\midrule
per-token dynamic INT4     & 5.39 & 6.21 & 7.22 \\
persistent static INT4     & 4.82 & 4.08 & \textbf{2.55} \\
\midrule
\emph{zero-state control}  & \emph{3.05} & \emph{4.08} & \emph{3.19} \\
\midrule
persistent group-wise INT4 & \textbf{2.05} & \textbf{2.15} & 2.70 \\
\bottomrule
\end{tabular}
\caption{Which containment wins is architecture-dependent. The control sets a threshold: any policy above it
does worse than throwing the state away, and per-token dynamic INT4 is above it in all three models. Group-wise scaling earns its advantage on the
two delta-rule models and loses it on the SSM, where one persistent scale already clears the threshold; it loses
there while spending $5.0$ modeled bits per element against $4.0$ for the per-channel policies. Independent
held-out articles, $n=16$, natural continuation. The GDN
columns are the independent-article runs of \S\ref{sec:mech}, so dynamic INT4 reads $5.39$ here and $5.47$ in
Table~\ref{tab:contain}.}
\label{tab:cross}
\end{table*}

%% file: sec_policy.tex
\section{Policy $\times$ exposure}
\label{sec:policy}

\begin{figure*}[t]
\centering
\includegraphics[width=\textwidth]{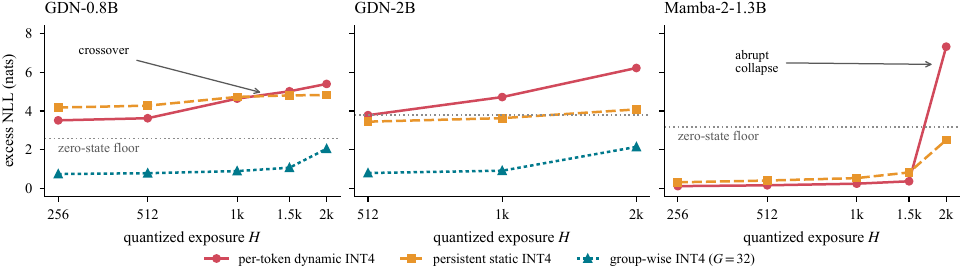}
\caption{\textbf{Policy $\times$ exposure.} Excess NLL against quantized exposure $H$, 16 independent documents
per point, total context fixed at 2048 so every $H$ predicts the same continuation. The dynamic$-$static gap
widens with $H$ in all three models; whether dynamic is outright better at short $H$ is model-dependent. Dotted
line: zero-state control floor.}
\label{fig:policy}
\end{figure*}

If the damage accumulates through the recurrence, then which policy is best cannot be a fixed answer: it has to
depend on how long the state has been quantized. Using the matched-target continuation protocol of \S\ref{sec:setup} on
independent articles, Figure~\ref{fig:policy} reports the excess NLL of per-token dynamic, persistent static, and group-wise
INT4 across models.

\paragraph{The adaptive policy is the one that decays.} The policy that tracks the distribution most faithfully
is also the one whose disadvantage grows with every step of exposure: for every model, the dynamic$-$static
excess-NLL gap widens monotonically. We measure the interaction as a difference-in-differences, the growth in
that gap between $H=512$ and $H=2048$, and it is positive and significant in all three models. GDN-0.8B
widens by $1.22$ nats ($t=6.9$) and GDN-2B by $1.80$ ($t=7.3$); Mamba-1.3B widens by $5.06$ ($t=12.9$), and
an independent two-point run on that model replicates the effect at $4.83$ ($t=14.0$). Persistent calibration
therefore becomes more valuable, and per-token dynamic scaling more dangerous, precisely in the long-context
regime these architectures are built for.

\paragraph{And at short exposure it often looks like the better choice.} In two of our three models dynamic is
outright ahead at $H=512$, for
GDN-0.8B ($t=-5.32$) and Mamba-1.3B ($t=-13.81$), so a benchmark run at that length would select it. The
five-point curve of Figure~\ref{fig:policy} puts the GDN-0.8B crossover between $H=1024$, where dynamic is
still marginally ahead, and $H=1536$, where static has taken the lead. GDN-2B never shows the reversal. Static wins there at every exposure, including $H=512$ ($t=+2.64$), so the
gap only widens monotonically. Mamba-2 reverses abruptly instead,
from $+0.46$ nats at $H=1536$ to $-4.81$ at $H=2048$. The claim we are entitled to is
therefore the narrower one. \emph{Per-token dynamic scaling degrades relative to persistent calibration
monotonically with quantized exposure across architectures and scales, and whether it also loses outright at
short exposure depends on the model.}

\iflcfm
\else
\paragraph{Decomposition.} A history-only variant, which stops quantizing during the scored continuation, shows
no crossover at all. Dynamic's corrupted history is worse at every exposure, by $0.45$, $1.66$, and $2.95$ nats
at $H=512$, $1024$, and $2048$. Dynamic's short-exposure advantage in the full pipeline therefore comes from
its lower cost during the continuation stage, not from better history retention, and the crossover is the point
at which its accumulating history damage overtakes that lower per-step cost.

Static pays a higher but roughly stable per-step cost and accumulates little, its history damage growing only
from $0.39$ to $0.99$ over the same range. Dynamic pays little per step and accumulates steeply, from $0.84$ to
$3.94$. That split is what containment predicts. We do not treat the full and history-only losses as strictly
additive, because the two branches follow different state trajectories during the scored continuation.
\fi

%% file: sec_whatworks.tex
\section{Characterizing the residual}
\label{sec:whatworks}

Containment explains the exposure ranking of \S\ref{sec:policy}, but not the residual that yoking left behind.
To ask whether that residual depends on resolution, we repeat the yoking experiment from INT3 to INT8
(Table~\ref{tab:bits}).

\begin{table*}[t]
\centering\small\setlength{\tabcolsep}{6pt}
\begin{tabular}{lccccc}
\toprule
Scale yoked, $H=2048$ & INT3 & INT4 & INT5 & INT6 & INT8 \\
\midrule
excess NLL & 4.82 & 3.57 & 2.96 & 1.56 & \textbf{0.31} \\
\bottomrule
\end{tabular}
\caption{The residual thins with bit-width, which an accumulation floor would not do. GDN-0.8B, 16 held-out
articles, scale yoked throughout so the feedback cannot confound the sweep. The INT4 arms reproduce the $5.39$
self-scaled and $3.57$ yoked of \S\ref{sec:mech} to four decimal places.}
\label{tab:bits}
\end{table*}

The residual falls monotonically in 15 of the 16 articles, and $8.7\%$ of it survives at INT8, where the paired
gap to INT4 is $3.26$ nats ($t=32.5$). The residual is \emph{resolution-dependent}: more bits buy it back, and
none of the rearrangements of the same bits we test below does. This does not separate instantaneous
quantization loss from accumulation, since updates lost to a coarse grid accumulate coherently and that
accumulation also shrinks as the grid refines \citep{li2026preserving}.

\paragraph{Stochastic rounding is not the remedy.} Round-to-nearest is biased, and bias is what compounds
through a recurrence, so stochastic rounding \citep{gupta2015stochastic} is the standard fix and the one the
Nemotron~3 line ships for its SSM cache \citep{nemotron3ultra}. It costs $6.71$ nats on top of yoked INT4 ($t=41.1$) and $5.77$ on top of
self-scaled INT4 ($t=17.2$), falling to $0.62$ at INT8 ($t=10.5$). Removing the bias makes each step's error
larger---$\sqrt{2}$ in RMS, $1.42\times$ measured on the states---and in this setting the larger per-step error
outweighs the removed bias. Rounding biased toward the direction of each update is a different intervention,
and can help even as it enlarges the per-step error \citep{li2026preserving}. Against exposure it holds all three policies between $9.7$ and $11.2$ nats out to 32K, worse at every
checkpoint than clearing the state (Appendix~\ref{sec:appendix-residual}).

\paragraph{Three ways of spending the same bits.} Each of the following gives the quantizer more information
than uniform group-wise scaling does, after the outlier-aware designs that work on weights, activations and the
KV cache \citep{lin2023awq,hooper2024kvquant}, and each is beaten by it.
\begin{itemize}[leftmargin=*,itemsep=1pt,topsep=1pt]
\item \emph{Outlier-frequency channel targeting / exact sparse escape.} Giving extra bits to the most
frequently-overflowing channels, or storing overflowing elements exactly, does not beat a random channel subset
or uniform allocation---locating outliers is not containing the feedback.
\item \emph{Dynamic$\to$static and static$\to$dynamic hybrids.} A hybrid that runs dynamic early and static
late does beat \emph{both} pure policies, because the late static caps the runaway. It still loses to group-wise
($t=-14.2$).
\item \emph{Capped dynamic scaling.} Beats static, loses to group-wise ($t=-10.0$). Its exact-escape variant
adds nothing (Table~\ref{tab:contain}): the cap is what helps, not the escape.
\end{itemize}

\paragraph{Uniform group-wise scaling beats all three.} Persistent group-wise INT4 is the least-damaging policy at
every GDN model and horizon, at a lower modeled bit rate than uniform INT5. It beats persistent static
INT4 by $t=-25$ to $-29$ at 0.8B and $t=-12$ at 2B, and it beats each design above by $t=-10$ to
$-19$ (Table~\ref{tab:contain}). Static and group-wise
both remove the feedback edge and differ only in how many scales they spend on the channel axis, whereas every
design above spends its extra information on \emph{finding} outliers instead. This is the least damaging uniform
policy on these models, and we did not invent any part of it.

\iflcfm
\else
\paragraph{A caution for practitioners.} Even the best policy is far from lossless. At $H=2048$ uniform
group-wise INT4 leaves natural-text perplexity at $7.8\times$ the unquantized model on 0.8B and $8.6\times$ on 2B,
and the best leak-free allocation still costs $4.9\times$ and $7.4\times$. Both pure INT4 policies are worse here than
clearing the state altogether, so none of these options is usable. \emph{Quantization that preserves perfect
NIAH recall can still severely damage general language-modeling quality}: recall of a single planted fact is a
far weaker probe of state fidelity than natural continuation.
\fi

\iflcfm
\else

\begin{figure*}[t]
\centering
\includegraphics[width=\textwidth]{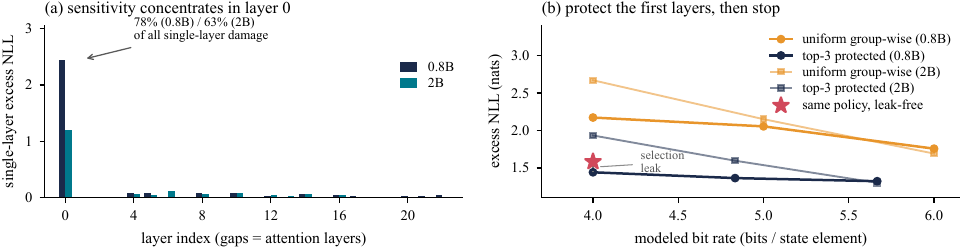}
\caption{\textbf{Per-layer allocation.} (a) Quantizing one layer at a time, plotted at the true layer index;
gaps are attention rather than linear-attention layers. (b) Rate--distortion. The connected series are an
\emph{in-sample} sweep, optimistic in level because the protected layers were chosen on the articles they are
scored on. The star re-measures the same 4.0-bit policy under the leak-free protocol of
Table~\ref{tab:alloc}, and that gap is why the curves sit below the table.}
\label{fig:alloc}
\end{figure*}

\paragraph{Does per-layer mixed precision help? A leak-free test.} Per-layer mixed precision is the natural
refinement, because single-layer sensitivity is strongly concentrated. Quantizing one layer at a time to INT4,
the first GDN layer alone accounts for $63$--$78\%$ of the summed single-layer damage at $H=2048$
(Figure~\ref{fig:alloc}a). The obvious move is to protect a few layers at high precision and quantize the rest
aggressively.

We test that under a leak-free protocol, with calibration, selection, and test articles disjoint by title hash
(6, 15, and 18 articles). Layer selection runs under the \emph{deployment} quantizer, persistent group-wise, by
marginal rescue from an all-INT2 baseline, and every reported number is on the held-out test articles
(Table~\ref{tab:alloc}, Figure~\ref{fig:alloc}b).

\emph{The positive finding.} Extreme allocation---top-3 layers at INT8, the rest at INT2, averaging 4.0
bits---is a genuine Pareto improvement over uniform group-wise at a matched budget, beating uniform INT3 at the
same 4.0 bits by $t=-12.8$ at 0.8B. It also beats uniform INT4 at a 20\% lower rate, though that higher-rate
margin attenuates with model scale, from $t=-10.3$ at 0.8B to $t=-2.2$ at 2B. Choosing the right layers
is what does the work: the rescue-selected allocation beats random, inverse, and reconstruction-MSE-ranked
allocations by $t=-17$ to $-22$.

\emph{The deflationary one.} The sensitive layers are simply the \emph{first} layers. Rescue selection, a
matched-budget knapsack optimizer, and a trivial ``protect the first three layers'' baseline return numerically
identical results, not merely similar ones (Table~\ref{tab:alloc}). No learned or search-based allocation adds
anything over the trivial heuristic. Early recurrent layers dominate
quantization sensitivity and protecting them helps at a fixed budget, but no sophisticated allocation is
warranted, and even the first-layer advantage over spending the bits uniformly shrinks as models grow.

\begin{table}[t]
\centering\small\setlength{\tabcolsep}{3.6pt}
\begin{tabular}{lcc}
\toprule
Allocation (avg 4.0 bits, $H=2048$) & 0.8B & 2B \\
\midrule
\textbf{extreme (top-3 INT8, rest INT2)} & \textbf{1.58} & \textbf{2.00} \\
knapsack (matched-budget optimizer) & 1.58 & 2.00 \\
first-3-layers (trivial baseline)   & 1.58 & 2.00 \\
uniform INT3                        & 2.35 & 2.52 \\
uniform INT4 (5.0 bits)             & 2.10 & 2.17 \\
random 3 layers                     & 3.1--3.2 & 3.3 \\
reconstruction-MSE top-3            & 3.24 & 3.43 \\
inverse (least-sensitive 3)         & 3.34 & 3.61 \\
\bottomrule
\end{tabular}
\caption{Leak-free per-layer allocation on 18 held-out test articles (excess NLL). Protecting the right layers
helps (vs.\ random/inverse/MSE), but the ``right'' layers are just the first ones: extreme, a matched-budget
knapsack, and a trivial protect-first-three baseline are \emph{identical}. Sophisticated selection adds nothing.}
\label{tab:alloc}
\end{table}
\fi

%% file: sec_survival.tex
\begin{figure*}[t]
\centering
\includegraphics[width=\textwidth]{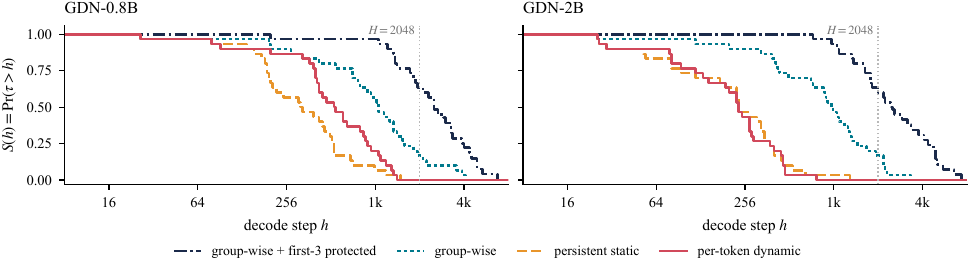}
\caption{\textbf{Survival of confident-token faithfulness.} $S(h)$ is the probability that a quantized model has
not yet flipped an argmax the FP model predicted with top-2 margin above 8, over 30 held-out articles,
right-censored at article length. The ordering is monotone at both scales, yet the best policy looks lossless at
$h\le 512$ and has failed on a third of articles by $h=2048$.}
\label{fig:survival}
\end{figure*}

\section{How long is a quantized state safe?}
\label{sec:survival}

Excess NLL averages over a window and hides \emph{when} a policy first breaks, so we add a reliability view.
Feeding the FP and a quantized model the same tokens, we record the first step at which the quantized model flips
a next-token argmax that FP had predicted with a top-2 logit margin above 8. That first-passage failure time
$\tau$ is measured on 30 held-out articles up to $H=8$K, right-censored at article length.

The survival curve $\Pr(\tau>h)$ (Figure~\ref{fig:survival}) orders the policies monotonically at every
horizon, and at every FP-confidence threshold (Appendix~\ref{sec:appendix-margin}): first-layer-protected
group-wise, then group-wise, then static and dynamic together. At $H=2048$,
protected group-wise leaves 19 of 30 articles confident-token-faithful, group-wise 5 of 30, and the two pure
policies none.

\textbf{There is no strategy reversal here.} The best policy short is the best policy long, so on this metric
short evaluation does not select the wrong policy, unlike the crossover of \S\ref{sec:policy}.

\textbf{But short evaluation badly underestimates risk.} The deployment-preferred policy looks essentially
lossless at short horizon, failing on 1 of 30 articles by $H=512$. By $H=2048$ it has failed on 11 of 30, and
those long-horizon failures have no short-horizon precursor. It replicates at 2B: 0 of 30 by
$H=512$, 12 of 30 by $H=2048$. Failure horizons cannot be
extrapolated from short traces either: a constant- or power-law-hazard fit to the first 256 steps mis-predicts
survival at $H=2048$ by well over an octave, because the good policy has no short-window events to fit.
Evaluating recurrent-state quantization at short length is hazardous not because it ranks policies wrongly, but
because it certifies a policy as safe long before its failures appear.

%% file: sec_discussion.tex
\section{Discussion and conclusion}
\label{sec:cross}

\paragraph{What generalizes.} A single-architecture result would not be worth much, so it matters that the
exposure interaction (\S\ref{sec:policy}) and the failure frontier both replicate across GDN (0.8B, 2B) and
Mamba-2 (1.3B), the frontier sitting about twice as far out on the SSM. Group-wise dominance does not: it holds
at both GDN scales and fails on Mamba-2 (Table~\ref{tab:cross}). Concurrent work approaches
low-precision recurrent inference by other routes---decay-aware mixed precision applied to the state itself
\citep{damp}, and a per-element floating-point format applied to weights and activations with the state left in
high precision \citep{minima}---neither of which we compare against here. Severity varies with
architecture while the functional interaction does not, so the interaction belongs to quantized recurrent-state
decoding and not to the delta rule.

\paragraph{One principle, four findings.} The positive results are one idea seen four ways: a persistent scale
\emph{eliminates} the feedback edge, a calibrated ceiling \emph{bounds} it, periodic reset \emph{interrupts} it,
and a higher bit-width merely delays the loop. That every outlier-locating policy we tried failed under per-step quantization is the
same principle read backwards.

\paragraph{What we would tell a practitioner.} Prefer any persistently calibrated scale over per-token dynamic
scaling, check group-wise against plain static on the target architecture rather than assuming group-wise wins,
protect the first recurrent layer, and expect little from learned allocation beyond that. The point we
would most want carried forward is the reporting one: a bit-width certified at 512 steps says little about 4K and nothing about 32K,
and the horizon is where low-bit recurrent-state inference actually breaks.

%% file: sec_backmatter.tex
\iflcfm
\else

\section*{Limitations}
\label{sec:limitations}

\textbf{Measurement.} We simulate quantization rather than implement a packed kernel, so bit rates are modeled
and we report no latency or throughput. The best policy still leaves perplexity at $7.8\times$ (0.8B) and
$8.6\times$ (2B) the unquantized model at $H=2048$. We run no head-to-head against architecture-specific quantizers
(Q-Mamba, Quamba2, DAMP) or against concurrent recurrent-state quantizers
\citep{li2026preserving,chen2026lowbit,pan2026leapquant}, which differ in both what they quantize and at what
rate, and make no systems claim.

\textbf{Scope and mechanism.} Two architectures and three model sizes, all below 3B; the first-layer-protection
advantage already shrinks from 0.8B to 2B, so we cannot say whether it survives at deployment scale. Scale-yoking
identifies the feedback as the cause of the \emph{numerical} runaway but recovers only part of the functional
loss, and we do not identify the remainder; coherent accumulation of rounding-lost updates
\citep{li2026preserving} is a candidate our bit-width sweep cannot separate from instantaneous loss.
\S\ref{sec:mech} records the countervailing evidence and the run-to-run instability of the norm itself.

\section*{Ethics statement}
\input{sec_ethics}

\section*{Reproducibility}
\input{sec_repro}
\fi

%% file: sec_ethics.tex
This work is an empirical analysis of numerical stability in quantized inference. It uses publicly released model
checkpoints and the public WikiText-103 corpus, introduces no new data collection, and involves no human subjects
or personally identifying information. The synthetic recall probes are generated from fictional entities. The
practical effect of the findings, if any, is to make low-bit long-context inference more reliable and to warn
against a specific evaluation practice (certifying a quantizer at short decode length) that can overstate
deployed reliability. We see no direct dual-use concern beyond those already attached to efficient LLM inference
generally.

%% file: sec_repro.tex
All experiments use public model checkpoints and the WikiText-103 corpus, and every run reported here fits on a
single RTX 3090. Code, result files and the project page are at
\url{https://avalon-s.github.io/dynamic-scaling-fails/}. The release carries the fake-quantization, calibration and
evaluation code, the article-disjoint calibration/selection/test splits by title hash, the result file behind
every number reported here, the plotting scripts that build every figure directly from those files, and the
per-run provenance (article IDs, quantizer, group size, model and data revision).

GDN runs are bit-identical between independent executions: re-running the
long-horizon sweep of Appendix~\ref{sec:appendix-32k} from scratch reproduced all 35 per-document excess-NLL
values to the last digit. The Mamba-2 runtime is not deterministic, and repeating the same configuration moves a
single-document loss by a few hundredths of a nat, so Mamba numbers should be read as reproducible in
distribution rather than exactly. State norms reproduce far less well than losses, because a runaway amplifies
whatever differs between runs: across three executions the dynamic-INT4 norm at $H=2048$ ranges from $38.9$ to
$42.3$ while the corresponding excess NLL moves by $0.4\%$. Read a norm from a runaway condition to one
significant figure.

%% file: sec_appendix.tex
\section{The failure frontier in numbers}
\label{sec:appendix-frontier}

Table~\ref{tab:frontier} gives the exact recall values plotted in Figure~\ref{fig:frontier}a.

\begin{table}[h]
\centering\small\setlength{\tabcolsep}{3.0pt}
\begin{tabular}{llccccc}
\toprule
Model & bits & $H=0$ & 1024 & 2048 & 4096 & 8192 \\
\midrule
\multirow{4}{*}{GDN-0.8B}
 & INT8 & 1.00 & 1.00 & 1.00 & 1.00 & 1.00 \\
 & INT5 & 1.00 & 0.92 & 0.67 & 0.33 & 0.00 \\
 & INT4 & 1.00 & 0.08 & 0.00 & 0.00 & 0.00 \\
 & INT3 & 1.00 & 0.00 & 0.00 & 0.00 & 0.00 \\
\midrule
\multirow{3}{*}{GDN-2B}
 & INT8 & 1.00 & 1.00 & 1.00 & 1.00 & --- \\
 & INT5 & 1.00 & 0.83 & 0.08 & 0.00 & --- \\
 & INT4 & 1.00 & 0.17 & 0.00 & 0.00 & --- \\
\midrule
\multirow{5}{*}{Mamba-2-1.3B}
 & none & 1.00 & 1.00 & 1.00 & 0.75 & 0.00 \\
 & INT8 & 1.00 & 1.00 & 1.00 & 0.08 & 0.00 \\
 & INT5 & 1.00 & 0.92 & 0.67 & 0.00 & 0.00 \\
 & INT4 & 1.00 & 1.00 & 0.33 & 0.00 & 0.00 \\
 & INT3 & 0.92 & 0.08 & 0.00 & 0.00 & 0.00 \\
\bottomrule
\end{tabular}
\caption{NIAH recall (fraction, 12 seeds) by bit-width and quantized decode horizon $H$. The unquantized GDN
models score 1.00 throughout. On Mamba-2 the probe expires with horizon, so its unquantized row is given as
well; at $H=8192$ no condition answers and the column carries no information. The Mamba-2 INT5 and INT4 rows
differ by one trial at $H=1024$, which is the resolution of 12 seeds.}
\label{tab:frontier}
\end{table}

\section{Calibration robustness of persistent scaling}
\label{sec:appendix-calib}

How the calibration set is chosen matters little for persistent scaling (Table~\ref{tab:calib}, NIAH recall
probe). Mis-scaling the calibrated step size, calibrating on a different sequence, or
using a short calibration set all leave persistent recall far above per-token dynamic INT4, which is at $0.00$ by
$H=2048$ in every setting. The scale has to be \emph{decoupled} from the instantaneous state; it does not have
to be tuned precisely.

\begin{table}[h]
\centering\small
\begin{tabular}{lcc}
\toprule
Calibration setting & $H=1024$ & $H=2048$ \\
\midrule
persistent, held-out (default) & 0.88 & 0.88 \\
\;\;step size $\times0.5$      & 1.00 & 1.00 \\
\;\;step size $\times2.0$      & 0.69 & 0.44 \\
\;\;different sequence            & 0.94 & 0.62 \\
\;\;short calibration set         & 1.00 & 0.81 \\
\midrule
per-token dynamic (any)          & 0.00 & 0.00 \\
\bottomrule
\end{tabular}
\caption{NIAH recall of persistent INT4 under calibration perturbations (GDN-0.8B, 16 seeds). Only $2\times$
over-scaling, which wastes resolution the way dynamic does, degrades it materially.}
\label{tab:calib}
\end{table}

\section{Additional protocol notes}
\label{sec:appendix-proto}
\textbf{Zero-state control.} For every task, clearing the recurrent state each step (rather than quantizing it)
collapses performance to the no-context floor (excess NLL $2.3$--$4.1$ across models and horizons; the dotted
line in Figure~\ref{fig:policy}), confirming the tasks genuinely exercise the recurrent state.
\textbf{Boundary reset.} The periodic-reset result (Table~\ref{tab:contain}) zeros the state on a fixed schedule.
Its benefit is insensitive to the period: excess NLL is $3.38$, $3.68$, and $3.44$ for
$K\in\{128,256,512\}$. A tuned corruption schedule would depend on the period; interrupting the accumulation
does not. \textbf{What the state is worth in memory.} The recurrent state is fixed in size. An attention KV cache is not, so
the share of decode memory the state accounts for falls as context grows. Qwen3.5-0.8B carries 18 linear layers
of $16\times128\times128$ FP32 state, $18$\,MB in all, against six attention layers with two KV heads of
dimension 256 in BF16, $12$\,KB per token. That is $43\%$ of the cache at 2K, $16\%$ at 8K, $4.5\%$ at 32K and
$1.2\%$ at 128K. Mamba-2-1.3B has no attention layers at all, so its $96$\,MB of state is the whole cache at
every length. The case for quantizing the state is strongest in pure SSMs and at moderate context. What this
paper measures is the cost of doing it.
\textbf{State-norm provenance.} The two self-scaled means quoted in \S\ref{sec:mech} come from different runs
on disjoint articles, $3.9\times10^{3}$ from the exposure sweep and $3.5\times10^{4}$ from the yoking run. The
distribution is heavy-tailed. On the yoking run 5 of 16 articles exceed $10^{3}$, and the median is $268$
against that mean. Excess NLL for the same condition moves by under $2\%$ across the same runs, so the
functional measurements are stable where the norm is not.
\textbf{Multi-key retrieval and per-model exposure curves} are reported with the released code.

\section{What the residual is made of}
\label{sec:appendix-residual}

\S\ref{sec:whatworks} shows that the residual surviving yoking is resolution-dependent, on the evidence of a
bit-width sweep. This section carries the
exposure half of that comparison, which does not fit there, and the protocol both probes share. All of it runs
on the 16 held-out articles of \S\ref{sec:mech}.

\paragraph{Protocol.} Both probes hold the scale yoked. The bit-width sweep and the stochastic-rounding
penalties are in \S\ref{sec:whatworks}; stochastic rounding replaces the deterministic round with
$\lfloor x/s + u \rfloor$, $u \sim U(0,1)$.

Against exposure, on the seven long contexts of Appendix~\ref{sec:appendix-32k}, the effect is larger.
The unquantized reference is bit-identical across the two jobs at every checkpoint, so the policies pair by
document. Stochastic rounding holds all three policies between $9.7$ and $11.2$ nats at every horizon from 2K to
32K, where round-to-nearest separates them by $2.0$ to $5.9$. The distinctions the rest of the paper draws
mostly vanish. Persistent static and group-wise differ by $2.76$ nats under round-to-nearest at $H=32768$
($t=9.3$) and by $0.54$ under stochastic rounding ($t=1.6$). It also leaves the model worse off than clearing
the state entirely, exceeding the zero-state control by $6.6$ to $7.1$ nats at every checkpoint ($t=19$ to
$32$).

\section{The curve past 2048}
\label{sec:appendix-32k}

Fixing total context at 2048 is what lets every $H$ predict the same continuation, and it is also a ceiling. It
cannot be lifted by running longer: the held-out corpus contains no article longer than $16{,}227$ tokens, so a
32K single-article context does not exist to be measured.

This appendix uses a second protocol. Each context is quantized from the first decode step, and a 256-token
continuation is scored at $H\in\{2048, 4096, 8192, 16384, 32768\}$. Excess NLL is still the quantized minus
unquantized loss on the same tokens at the same position, so the drift of natural perplexity with position
cancels in the difference rather than entering it. Seven contexts of $33{,}024$ tokens are built by concatenating
held-out articles; no article appears in two contexts, nor in the six articles used for calibration.

The two protocols coincide at $H=2048$, and that overlap is the check. This run gives $5.89$, $4.51$, and $2.02$
nats for per-token dynamic, persistent static, and group-wise INT4, against the $5.39$, $4.82$, and $2.05$ of
Figure~\ref{fig:policy}. Each published value falls inside the new mean plus or minus two standard errors over
the seven contexts, and the ordering is unchanged.

\begin{figure*}[t]
\centering
\includegraphics[width=\textwidth]{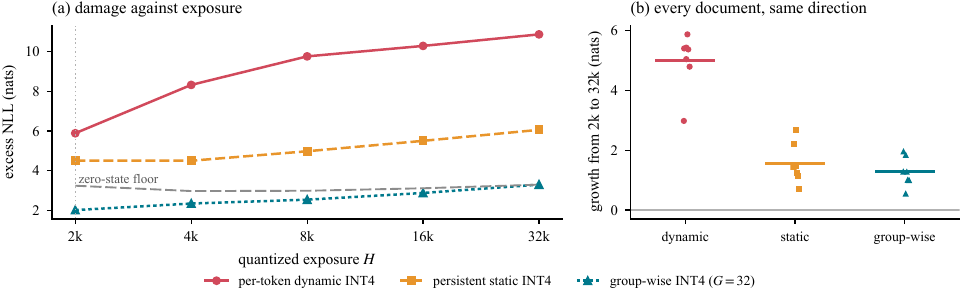}
\caption{\textbf{The curve past 2048.} (a) Excess NLL under the from-the-first-step protocol, seven held-out
contexts; the dotted vertical marks the longest exposure the matched-target design reaches. (b) The same
comparison per context, since the claim is a paired one. Bars are means.}
\label{fig:h32k}
\end{figure*}

Damage keeps accumulating past that point for every policy. Paired across the seven contexts, excess NLL grows
from $H=2048$ to $H=32768$ by $4.98$ nats for dynamic ($t=13.9$), $1.56$ for static ($t=6.2$), and $1.28$ for
group-wise ($t=6.9$), and no context moves the other way in any policy. Group-wise INT4 remains the least
damaging at every checkpoint, by $t=17.0$ against dynamic and $t=9.3$ against static at $H=32768$. In the units
of \S\ref{sec:whatworks} its perplexity penalty grows from $7.5\times$ the unquantized model at $H=2048$ to
$27\times$ at $H=32768$.

\paragraph{How long the state stays worth keeping.} Excess NLL says how much a policy costs, not whether the
state it maintains is still worth maintaining. Running the same seven contexts with the recurrent state zeroed at
every step answers that directly. Group-wise INT4 beats the zeroed state by $1.23$ nats at $H=2048$
($t=-7.4$), and the margin decays with exposure: $0.63$ at $H=4096$, $0.45$ at $H=8192$, $0.24$ at
$H=16384$, and $0.00$ at $H=32768$ ($t=0.0$). Past roughly 16K the difference stops being detectable at this
sample size, and at 32K the quantized state carries no more of the context forward than discarding it would. The
zeroed control itself is flat across horizon, at $2.98$ to $3.30$ nats, inside the $2.3$--$4.1$ band this control
gives elsewhere in the paper.

This also settles how to read a curve that flattens. Per-token dynamic sits three times the floor at $H=32768$
and has been above it since $H=2048$, so its deceleration after 8K is exhaustion rather than containment.
Group-wise flattens nowhere in the range, and converges on the floor from below instead.

\paragraph{What ``safe'' means for INT8.} The frontier of \S\ref{sec:frontier} calls INT8 safe at every horizon
it measures, on the evidence of a recall probe that returns $1.00$ throughout. Natural continuation disagrees.
Under the same protocol, group-wise INT8 costs $0.26$ nats at $H=2048$ and $0.62$ at $H=32768$, growing by
$0.35$ paired across contexts ($t=10.9$); its perplexity penalty is $1.30\times$ the unquantized model at the
short horizon and $1.85\times$ at the long one. Group-wise INT5 sits between, at $1.84$ and $3.11$ nats.

\paragraph{Scale and architecture.} The growth replicates at 2B and more strongly: group-wise INT4 adds $1.70$
nats between $H=2048$ and $H=32768$ ($t=8.0$), static $1.42$ ($t=4.5$). Per-token dynamic diverges out of
bfloat16 range on one of the seven contexts before $H=16384$, so its 2B growth of $3.06$ nats ($t=5.5$) is over
the six that remain finite; the numerical runaway of \S\ref{sec:mech} becomes an actual overflow at this scale.
Mamba-2-1.3B does \emph{not} replicate it. Over the same seven contexts the SSM state shows no significant growth
past $H=2048$ under either policy, $-0.26$ nats for dynamic ($t=-0.5$) and $+0.68$ for static ($t=1.9$),
though dynamic is already $7.06$ nats behind at $H=2048$ and static remains ahead of it by $t=9.2$ at
$H=32768$. Whether the long-horizon accumulation is specific to the delta-rule family or harder to
resolve against this runtime's nondeterminism is not settled by seven documents.

\section{Confidence threshold in the survival analysis}
\label{sec:appendix-margin}

The failure event in \S\ref{sec:survival} requires the full-precision model to have predicted with a top-2 logit
margin above $8$. A lower threshold counts near-ties as failures, so every time-to-failure shortens, but the
ordering of policies is unchanged (Table~\ref{tab:margin}). The threshold sets the scale of $\tau$, not the
ranking it induces.

\begin{table}[h]
\centering\small\setlength{\tabcolsep}{3pt}
\begin{tabular}{lccccc}
\toprule
Policy & $\mu=0$ & $1$ & $2$ & $4$ & $8$ \\
\midrule
\multicolumn{6}{l}{\emph{GDN-0.8B}} \\
group-wise + first 3           & 3 & 20 & 58 & 422 & 2538$^\dagger$ \\
group-wise                    & 3 & 14 & 27 & 111 & 1059 \\
dynamic                       & 2 & 12 & 22 & 64  & 542 \\
static                        & 2 & 9  & 14 & 34  & 322 \\
\midrule
\multicolumn{6}{l}{\emph{GDN-2B}} \\
group-wise + first 3           & 4 & 25 & 49 & 242 & 2502$^\dagger$ \\
group-wise                    & 3 & 14 & 25 & 105 & 987 \\
dynamic                       & 2 & 11 & 13 & 22  & 228 \\
static                        & 2 & 7  & 12 & 25  & 230 \\
\bottomrule
\end{tabular}
\caption{Median first-failure step $\tau$ over 30 held-out articles, by FP-confidence threshold $\mu$.
$^\dagger$ some articles right-censored at article length, so the true median is larger.}
\label{tab:margin}
\end{table}

